\documentclass{ws-rv961x669}
\usepackage{ws-rv-van}     % numbered citation/references (default)
\usepackage{ws-rv-thm}     % comment this line when `amsthm / theorem / ntheorem` package is used
\usepackage{subfigure}     
\usepackage{graphicx}
\usepackage{amsmath}
\usepackage{amssymb}
\usepackage{booktabs}
\usepackage{footmisc} 
\usepackage{multicol} 
\usepackage{array}
\usepackage{caption}

\newcommand{\mysection}[1]{\vspace{0.5em}\noindent\textbf{#1}}

\usepackage{hyperref}
\usepackage[normalem]{ulem}  % Enables underlining
\hypersetup{
    colorlinks=true,
    urlcolor=blue
}

\RequirePackage{xspace}
\makeatletter
\DeclareRobustCommand\onedot{\futurelet\@let@token\@onedot}
\def\@onedot{\ifx\@let@token.\else.\null\fi\xspace}

\def\eg{\emph{e.g}\onedot} 
\def\ie{\emph{i.e}\onedot} 
\def\etal{\emph{et al}\onedot}

\begin{document}

\newcommand{\videoset}{\mathcal{V}}
\newcommand{\video}[1]{v^{#1}}
\newcommand{\nvideos}{N}
\newcommand{\videoidx}{n}
\newcommand{\videolength}{L}

\newcommand{\image}[2]{i_{#1}^{#2}}
\newcommand{\nimages}[1]{J^{#1}}
\newcommand{\imageidx}{j}

\newcommand{\actionset}[1]{\mathcal{A}^{#1}}
\newcommand{\nactions}[1]{A^{#1}}
\newcommand{\action}[2]{a_{#1}^{#2}}
\newcommand{\actionidx}{k}
\newcommand{\class}{c}
\newcommand{\classset}{\mathcal{C}}
\newcommand{\nclasses}{C}
\newcommand{\timestamp}{t}
\newcommand{\interframe}{T}
\newcommand{\timestampIndex}{\tau}

\newcommand{\RMSNetURL}{\scriptsize\url{https://github.com/aimagelab/RMSNet_Soccer}}
\newcommand{\CALFURL}{\scriptsize\url{https://github.com/SoccerNet/sn-spotting/tree/main/Benchmarks/CALF}}
\newcommand{\NetVLADURL}{\scriptsize\url{https://github.com/SoccerNet/sn-spotting/tree/main/Benchmarks/Pooling}}

\newcommand{\ASTRAURL}{\scriptsize{\url{https://github.com/arturxe2/ASTRA}}}
\newcommand{\HCMUSPKURL}{\scriptsize{\url{https://github.com/Fsoft-AIC/UGLF}}}
\newcommand{\COMEDIANURL}{\scriptsize{\url{https://github.com/juliendenize/eztorch}}}
\newcommand{\SoaresURL}{\scriptsize{\url{https://github.com/yahoo/spivak}}}
\newcommand{\ETWOESpotURL}{\scriptsize{\url{https://github.com/jhong93/spot}}}
\newcommand{\ZhuURL}{\scriptsize{\url{https://github.com/ArthurUnic/action_spotting}}}
\newcommand{\ZhouURL}{\scriptsize{\url{https://github.com/baidu-research/vidpress-sport}}}
\newcommand{\NetVLADPPURL}{\scriptsize{\url{https://github.com/SoccerNet/sn-spotting/tree/main/Benchmarks/TemporallyAwarePooling}}}
\newcommand{\CALFPlayerLocURL}{\scriptsize{\url{https://github.com/SoccerNet/sn-spotting/tree/main/Benchmarks/CALF_Calibration_GCN}}}

\newcommand{\BaikulovURL}{\scriptsize{\url{https://github.com/lRomul/ball-action-spotting}}}
\newcommand{\WangURL}{\scriptsize{\url{https://github.com/ZJLAB-AMMI/E2E-Spot-MBS}}}

\newcommand{\TDEEDURL}{\scriptsize{\url{https://github.com/arturxe2/T-DEED}}}

\global\long\def\method{\mathcal{M}\xspace}
\newcommand{\predicted}[1]{\widehat{#1}}
\newcommand{\confidence}{s}

\global\long\def\backbone{\mathcal{B}\xspace}
\global\long\def\neck{\mathcal{N}\xspace}
\global\long\def\head{\mathcal{H}\xspace}
\newcommand{\featureset}[1]{\mathcal{F}^{#1}}
\newcommand{\neckfeatureset}[1]{\widetilde{\mathcal{F}^{#1}}}
\newcommand{\feature}[2]{f_{#1}^{#2}}
\newcommand{\neckfeature}[2]{\feature{#1}{#2}'}

% Metric macros
\newcommand{\amap}{a\mbox{-}mAP}
\newcommand{\aap}{a\mbox{-}AP}
\newcommand{\precision}{p}
\newcommand{\recall}{r}
\newcommand{\prsubset}{\Phi}
\newcommand{\map}[1]{mAP@#1}
\newcommand{\ap}[2]{AP^{#1}_{#2}}
\newcommand{\timeinterval}{\delta}
\newcommand{\timeintervalstart}{\delta_{l}}
\newcommand{\timeintervalstop}{\delta_{h}}
\newcommand{\ntimeintervals}{\Delta}
\newcommand{\confidencethreshold}{\tau}
\newcommand{\truepositiveset}[3]{\mathcal{TP}^{#1}_{#2,#3}}
\newcommand{\ntruepositives}[3]{|\truepositiveset{#1}{#2}{#3}|}
\newcommand{\falsepositiveset}[3]{\mathcal{FP}^{#1}_{#2,#3}}
\newcommand{\nfalsepositives}[3]{|\falsepositiveset{#1}{#2}{#3}|}
\newcommand{\falsenegativeset}[3]{\mathcal{FN}^{#1}_{#2,#3}}
\newcommand{\nfalsenegatives}[3]{|\falsenegativeset{#1}{#2}{#3}|}
\newcommand{\precisionrecall}[2]{\mathcal{PR}^{#1}_{#2}}

\newcommand{\SNAS}{SoccerNet Action Spotting\xspace}
\newcommand{\SN}[1]{\SNAS (v#1)\xspace} % \SN{2} will display "SNAS (v2)" 
\newcommand{\SNBAS}{SoccerNet Ball Action Spotting\xspace}
\newcommand{\SNBASYear}[1]{\SNBAS (#1)\xspace} % \SNBASYear{2023} will display "SoccerNet Ball Action Spotting (2023)"

\chapter[Deep learning for action spotting in association football videos]{Deep learning for action spotting in association football videos}

\author[S. Giancola, A. Cioppa, B. Ghanem, and M. Van Droogenbroeck]{Silvio Giancola$^{1*}$, Anthony Cioppa$^{2*}$, Bernard Ghanem$^{1*}$, and \\Marc Van Droogenbroeck$^{2}$\footnote{All authors contributed equally.}}

\address{$^{1}$Center of Excellence for Generative AI, IVUL, KAUST, Saudi Arabia\\
$^{2}$Montefiore Institute, Open-SportsLab, University of Li{\`e}ge, Belgium}

\begin{abstract}

The task of \emph{action spotting} consists in both identifying actions and precisely localizing them in time with a single timestamp in long, untrimmed video streams.
Automatically extracting those actions is crucial for many sports applications, including sports analytics to produce extended statistics on game actions, coaching to provide support to video analysts, or fan engagement to automatically overlay content in the broadcast when specific actions occur.
However, before 2018, no large-scale datasets for action spotting in sports were publicly available, which impeded benchmarking action spotting methods.
In response, our team built the largest dataset and the most comprehensive benchmarks for sports video understanding, under the umbrella of \emph{SoccerNet}.
Particularly, our dataset contains a subset specifically dedicated to action spotting, called \emph{\SNAS}, containing more than $550$ complete broadcast games annotated with almost all types of actions that can occur in a football game.
This dataset is tailored to develop methods for automatic spotting of actions of interest, including deep learning approaches, by providing a large amount of manually annotated actions.
To engage with the scientific community, the SoccerNet initiative organizes yearly challenges, during which participants from all around the world compete to achieve state-of-the-art performances.
Thanks to our dataset and challenges, more than $60$ methods were developed or published over the past five years, improving on the first baselines and making action spotting a viable option for the sports industry.
This paper traces the history of action spotting in sports, from the creation of the task back in 2018, to the role it plays today in research and the sports industry.

\end{abstract}
\body

\begin{figure}[t]
    \centering
\includegraphics[width=\linewidth]{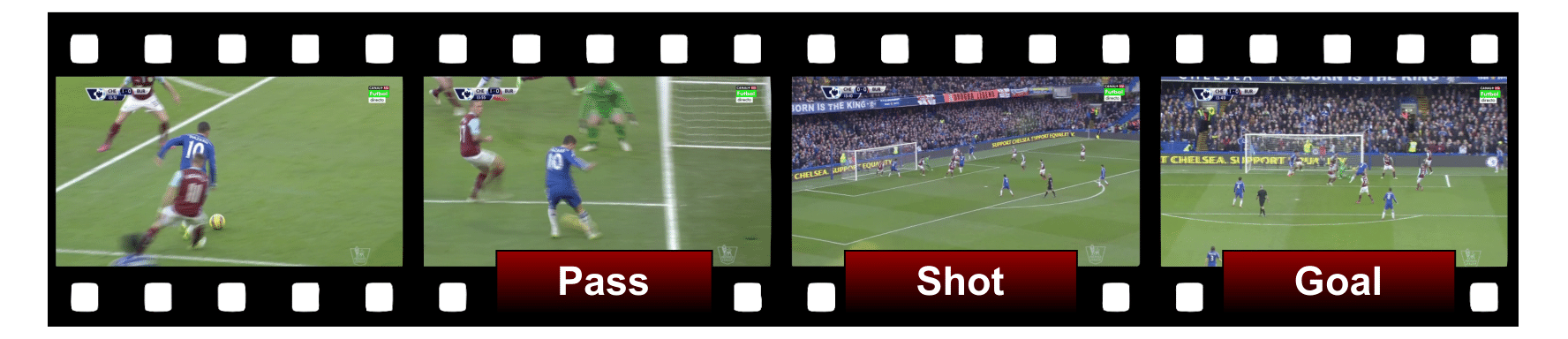}
    \caption{
    \textbf{Action spotting in football videos.}
    The task of \emph{action spotting} consists in both identifying actions and precisely localizing them in time with a single timestamp in long, untrimmed video streams.
    }
    \label{fig:SpottingPulling}
\end{figure}

\section{Introduction}

Sports video understanding has become an increasingly active research field~\cite{Moeslund2014Computer, Thomas2017Computer, Gadde2024TheComputer} as the sports industry continues to grow into a major global entertainment sector. 
In 2023, the global sports market generated over $160\$$ billion in annual revenue for sports equipment alone, while the betting part peaked at $242\$$ billion~\cite{GlobalSportsMarket}. 
Furthermore, new sectors such as eSport also open the potential for the market to grow even more in the coming years~\cite{GlobalESportsMarket}. 
Particularly, the rapid expansion in the volume of sports TV broadcasts has contributed to this demand, with the number of hours broadcasted in the United States alone growing more than fourfold between 2002 and 2017~\cite{LengthOfSports}. 
This surge in content has created a pressing need for automated analysis tools, as the manual review of video footage is both time-consuming and labor-intensive. 
Among various sports, football stands out as one of the most universally followed, with thousands of professional matches and millions of amateur games played annually. 
However, its dynamic and complex nature presents significant challenges for automatic analysis since much of the fine-grained understanding lies in subtle details. 
In response to these challenges, the scientific community has made remarkable strides over the past decade, tackling tasks such as player tracking~\cite{Cioppa2022SoccerNetTracking, Somers2024SoccerNetGameState} for individual performance evaluation, camera shot selection~\cite{Deliege2021SoccerNetv2} and dense video captioning~\cite{Mkhallati2023SoccerNetCaption} for enhancing broadcast production, and camera calibration~\cite{Cioppa2022Scaling, Magera2024AUniversal} for extracting physical metrics like speed and distance.

Action spotting focuses on identifying specific actions and precisely localizing them within long, untrimmed video streams using a single timestamp, marking the exact moment an action occurs~\cite{Giancola2018SoccerNet, Deliege2021SoccerNetv2, Wu2023ASurvey, Seweryn2023Survey-arxiv}, as illustrated in~\fref{fig:SpottingPulling}. 
This approach contrasts with traditional temporal video understanding~\cite{Caba2015ActivityNet}, where actions are defined by a start and end time. 
In sports, however, these boundaries can often be ambiguous. 
For instance, determining when a goal ``starts'' and ``ends'' is blurry. \emph{Does a goal begin with a pass or the shot itself? Does it end with the ball hitting the net or after the kick-off?} 
To address these ambiguities, Giancola~\etal~\cite{Giancola2018SoccerNet} introduced the novel task of action spotting in sports through the first version of the SoccerNet dataset. 
In this task, actions are annotated based on official football rules defined by the \emph{IFAB Laws of the Game}~\cite{IFAB2022Laws}, such as the goal being marked at the precise moment the ball crosses the line or a free-kick occurring when the player strikes the ball. 
This focus on precise localization differs from temporal activity localization, where defining the exact boundaries of activities is less critical~\cite{Caba2015ActivityNet}. 

Action spotting opens up a new area of research with a focus on precisely localizing actions, especially given the challenge of overlapping actions that may occur simultaneously. 
However, this task also brings inherent difficulties, including the sparsity of annotations since most moments in a match lack notable actions, and the discontinuities between adjacent frames that may be visually similar while representing different actions.

In sports, action spotting offers significant value across multiple verticals. 
First, it plays a crucial role in \emph{team strategy analysis}, by identifying key moments such as scoring opportunities, including corners or free kicks. 
This enables video analysts to efficiently extract and review historical data, gaining insights into how teams perform under specific circumstances. 
Second, action spotting contributes to \emph{fairer refereeing}, especially by localizing and characterizing potential foul moments. 
This provides referees and VAR officials with evidence to ensure consistency and fairness throughout matches~\cite{Held2023VARS, Held2024XVARS}. 
Third, action spotting \emph{enhances broadcasting experiences}, allowing broadcasters to personalize highlights based on a viewer's interests or time constraints~\cite{Valand2021AIBased, Valand2021Automated} 
 by automatically retrieving the actions of interest, or simply identifying the most salient moments.

Fourth, it benefits \emph{player scouting}, enabling teams to identify specific talents, such as a player skilled at scoring from headers during corner kicks. 
Automatically analyzing thousands of games therefore helps discover players with specialized abilities for particular sequences of actions. 
Lastly, sports \emph{medical analysis} benefits from spotting actions that could impact player health, such as headers~\cite{Giancola2023Towards}. 
By examining historical data, long-term effects of these actions can be evaluated. Likewise, action spotting can also help coaches and medical staff in real time by identifying potential health risks for players during games due to repeated dangerous actions.

Deep learning has demonstrated impressive performance across a wide range of tasks, from image classification to video understanding~\cite{Akan2022UseOfDeepLearning, Zhao2023ASurvey-arxiv, Pontes2024Application-ssrn, Pareek2020ASurvey}. 
However, these methods typically require large amounts of annotated data to train. 
To meet these data demands, deep learning models often rely on generic learning algorithms~\cite{Tan2019EfficientNet} trained on publicly available datasets~\cite{Deng2009ImageNet, Lin2014Microsoft} or fine-tuned algorithms trained on smaller, task-specific football datasets~\cite{Cioppa2018ABottomUp}. 
Before the introduction of SoccerNet~\cite{Giancola2018SoccerNet}, most datasets related to association football were either small, scattered, or private, making it difficult to fairly compare different methods and slowing the progress of scientific research in the field. 
Furthermore, many publicly available datasets suffered from size limitations, restricting their ability to generalize across different games~\cite{Homayounfar2017Sports, Biermann2021AUnified}.
The football research community, therefore, faced a pressing need for large-scale, publicly accessible datasets to address these challenges~\cite{Jiang2020SoccerDB,Feng2020SSET, VanZandycke2022DeepSportradarv1, Istasse2023DeepSportradarv2, Kassab2024TACDEC}. 
Nevertheless, creating such datasets is both time-consuming and expensive, as annotating large amounts of video data requires significant resources.

In response to the need for large-scale, publicly accessible datasets in sports video understanding, Giancola~\etal~\cite{Giancola2018SoccerNet} introduced SoccerNet in 2018. The goal was to create an open-source dataset for reproducible research and benchmarking in football video analysis. 
The original dataset comprised $500$ full broadcast football matches, totaling over $764$ hours of footage from the six major European leagues: Serie A, La Liga, Premier League, Ligue 1, Bundesliga, and Champions League, spanning from 2014 to 2017.
With this dataset, Giancola~\etal introduced the novel task of action spotting, focusing on the temporal localization of three key football actions: goals, cards, and substitutions. 
Over the years, SoccerNet has continuously expanded both in scope and content. 
SoccerNet-v2, presented by Deli{\`e}ge~\etal~\cite{Deliege2021SoccerNetv2} significantly increased the number of annotations and action classes, growing to include $110{,}458$ actions across $17$ classes, such as penalties, clearances, and ball out of play. 
The dataset also introduced annotations for camera shot changes, replay segments, and more, leading to new tasks like camera shot segmentation, boundary detection, and replay grounding.

The first SoccerNet challenge, organized in 2021, focused on action spotting and replay grounding, starting the dataset’s interest in the sports community. 
In 2022, SoccerNet-v3~\cite{Cioppa2022Scaling} and SoccerNet-Tracking~\cite{Cioppa2022SoccerNetTracking} were introduced, adding spatial annotations for players, the ball, and field elements across multiple views, alongside tasks such as pitch localization, camera calibration, player re-identification, and long-term multi-object tracking. 
In 2023, SoccerNet continued to expand with the addition of SoccerNet-Captions~\cite{Mkhallati2023SoccerNetCaption} for dense video captioning, SoccerNet-MVFouls~\cite{Held2023VARS} for multi-view foul recognition, and SoccerNet-BallActionSpotting~\cite{Cioppa2024SoccerNet2023Challenge} which identified first $2$, then $12$ different ball-related actions such as passes, drives, and headers. 
By 2024, further developments included SoccerNet-Depth~\cite{Leduc2024SoccerNetDepth} for monocular depth estimation, SoccerNet-XFoul~\cite{Held2024XVARS} for multi-modal question-answer triplets about refereeing decisions, and SoccerNet-GSR~\cite{Somers2024SoccerNetGameState} as the first open-source dataset for game state reconstruction.

Thanks to our dataset and challenges, more than $60$ methods have been developed or published over the past five years for action spotting alone, improving on the initial baselines by Giancola~\etal~\cite{Giancola2018SoccerNet} and Cioppa~\etal~\cite{Cioppa2020AContextaware}, and making action spotting a viable option for the sports industry. To improve accessibility of action spotting for researchers and practitioners, we also published an open-source library called \emph{OSL-Action Spotting}\cite{Benzakour2024OSL-ActionSpotting} that gathers several action spotting dataloaders, methods, and evaluation functions, under a common framework.

The remainder of this chapter is organized as follows. In ~\sref{sec:task}, we mathematically formalize the task of action spotting. 
\Sref{sec:data} provides a detailed description of the \SNAS datasets, explaining their evolution and the different versions dedicated to action spotting. 
\Sref{sec:methods} presents the various methods developed for action spotting, highlighting key approaches and advancements over the years.
In ~\sref{sec:metric}, we discuss the evaluation protocols and metrics used to assess the performance of action spotting methods. 
\Sref{sec:benchmark} offers a comprehensive benchmark, comparing the performance of state-of-the-art methods on the SoccerNet datasets. 
Finally, in \sref{sec:conclusion}, we conclude the chapter by summarizing the key findings for action spotting in both research and industry.

\section{Action Spotting: Definition of the Task}
\label{sec:task}

Action spotting is defined as the task of identifying and precisely localizing actions in time, \ie, with a single timestamp, in long, untrimmed video streams.
Given a set of $\nvideos$ untrimmed videos $\videoset = \{\video{1},\video{2},...,\video{\videoidx},...,\video{\nvideos}\}$, the objective is to extract the set of all actions $\actionset{\videoidx} = \{\action{1}{\videoidx},\action{2}{\videoidx},...,\action{\actionidx}{\videoidx},,...,\action{\nactions{\videoidx}}{\videoidx}\}$ for each video $\videoidx$, where $\nactions{\videoidx}$ is the total number of actions of interest in video $\videoidx$. All actions $\action{\actionidx}{\videoidx}$ are defined by two features: (1) a class $\class$ and (2) a timestamp $\timestamp$, indicating respectively the type of action and the exact moment the action happens in the video. Each class $c$ is chosen among a set $\classset$ of size $\nclasses$, corresponding to all possible classes of interest, while the timestamp is bounded by $\timestamp\in[0,\videolength]$, typically expressed in milliseconds. 
From a practical perspective, the videos are generally represented as a succession of frames $\video{\videoidx} = \{\image{1}{\videoidx},\image{2}{\videoidx}, ..., \image{\imageidx}{\videoidx},..., \image{\nimages{\videoidx}}{\videoidx}\}$, where $\nimages{\videoidx}$ is the number of frames in video $\videoidx$. This process discretizes the possible values of the timestamps of each action following: $\imageidx = \lfloor \frac{\timestamp}{\interframe} + \frac{1}{2} \rfloor + 1$, with $\interframe$ being the time between two frames, $\imageidx\in\{1,2,...,\nimages{\videoidx}\}$, and $\lfloor x + \frac{1}{2} \rfloor$ being the $round(x)$ function.

\section{SoccerNet: Large-Scale Datasets for Action Spotting}
\label{sec:data}

Recently, the evolution of sports video analysis has demanded large-scale, richly annotated datasets to fuel research and enable the development of robust data-driven models. Recognizing the lack of such resources, we introduced comprehensive datasets dedicated to action spotting in football videos. Since its inception, SoccerNet has become a cornerstone dataset for sports video understanding, growing in both scale and complexity over multiple iterations. \Tref{tab:datasets} provides a summary of the datasets released for action spotting under the SoccerNet umbrella.

\begin{table}[th]
\tbl{List of Action Spotting datasets published under the SoccerNet umbrella.\label{tab:datasets}}
{\begin{tabular}{@{}lcccr@{}}
\toprule 
\bf SoccerNet Dataset         & year & \# classes & \# games & \# annot.  \\ 
\colrule
\SN{1}~\cite{Giancola2018SoccerNet} & 2018 & ~3 & 500 & $6{,}637$ \\
\SN{2}~\cite{Deliege2021SoccerNetv2} & 2021 & 17 & 500+50 & $110{,}458$ \\
\SNBASYear{2023}~\cite{Cioppa2024SoccerNet2023Challenge} & 2023 & ~2 & ~~7+2~ & $11{,}041$   \\
\SNBASYear{2024}~\cite{Cioppa2024SoccerNet2024-arxiv} & 2024 & 12 & ~~7+2~ & $11{,}041$   \\
\botrule
\end{tabular}}    
\end{table}

\subsection{\SN{1}: the foundation}

The original release of \SN{1}, introduced by Giancola~\etal~\cite{Giancola2018SoccerNet} in 2018, was a pioneering dataset in the field of sports video analysis. The dataset contained $500$ full broadcast football matches, totaling $764$ hours of footage sourced from major European leagues like Serie A, Premier League, La Liga, Bundesliga, Ligue 1, and the UEFA Champions League.
\SN{1} primarily focused on three main action classes for the task of action spotting: \textit{Goals}, \textit{Cards}, and \textit{Substitutions}.
Each of these actions was annotated with a timestamp marking the moment the action occurred with a $1$-second precision. Different from traditional temporal activity localization tasks~\cite{Caba2015ActivityNet}, the actions are localized with a single timestamp. The simplicity of this approach allowed researchers to get their hands on developing methods for spotting actions in football games, laying the groundwork for more complex future developments.

\SN{1} became a cornerstone dataset and benchmark in the domain of sports video analysis, enabling the comparison of various methods under standardized conditions. The dataset also provided pre-extracted features from the video frames, based on ResNET~\cite{He2015DeepResidual}, C3D~\cite{Tran2015Learning} and I3D~\cite{Carreira2017QuoVadis}, to ease computational demands, which was crucial at the time due to hardware limitations.

\subsection{\SN{2}: increasing class diversity}

Building on the success of \SN{1}, \SN{2} was introduced by Deli{\`e}ge~\etal\cite{Deliege2021SoccerNetv2} in 2021. 
This version significantly expanded the dataset, both in terms of action classes and annotation complexity. Specifically, \SN{2} includes annotations for $17$ different action classes: \emph{Penalty, Kickoff, Goal, Substitution, Offside, Shot on target,  Shot off target, Clearance, Ball out of play, Throw-in, Foul, Indirect free-kick, Direct free-kick, Corner, Yellow card, Red card,} and \emph{Yellow to red card}.

The total number of annotated actions grew to $110{,}458$ across the dataset, now annotated with 1-frame precision, offering a more comprehensive representation of football events. In addition to increasing the variety of actions, \SN{2} introduced new tasks, such as \emph{replay grounding}, which consists in localizing the live action shown in a replay clip in a long untrimmed broadcast, and \emph{camera shot segmentation}, which consists in temporally segmenting the different camera shots during a broadcast. These additions opened up new avenues for research in other areas, such as broadcast production, automatic video summarization, and foul analysis~\cite{Held2023VARS,Held2024XVARS,Held2024Towards-arxiv}.

This dataset also served in the first \SNAS challenge in 2021, pushing the research community to develop state-of-the-art models and benchmark their methods for sports video analysis. Through this challenge, \SN{2} became a pivotal tool for advancing temporal spotting tasks in complex, real-world football scenarios.

\subsection{\SNBASYear{2023}: towards more precise localization}

In 2023, SoccerNet introduced the Ball Action Spotting task~\cite{Cioppa2024SoccerNet2023Challenge}. Unlike the broader action spotting task, which targets major game-changing events such as goals or cards, Ball Action Spotting requires models to precisely localize frequent interactions with the ball, offering a finer granularity of analysis. This task added a significant layer of complexity, as it required models to detect not just prominent actions, but also subtle and rapid ball exchanges that are critical to the game.

The Ball Action Spotting task was introduced during the SoccerNet challenges 2023~\cite{Cioppa2024SoccerNet2023Challenge} and only focuses on $2$ key actions: \emph{pass} defined as the moment the ball leaves the feet of a player, who is passing the ball, and \emph{drive}, defined as the moment the ball is received in the feet of another player, who is maintaining the control of the ball while moving through the field.
Though these actions are frequent and subtle compared to more prominent events like goals or fouls, they are essential for understanding team strategies, game dynamics, and individual player performance. 

The \SNBAS dataset comprises $7$ full games from the English football league with a total of $11{,}041$ annotated timestamps, providing data for training and validation. Additionally, $2$ extra games were held out as a separate evaluation set for the SoccerNet challenges 2023. This challenge pushed the research community to develop models capable of handling the increased temporal precision required for spotting ball actions, laying the groundwork for the future expansion of the task to incorporate more complex and diverse ball-related actions.

\subsection{\SNBASYear{2024}: precise localization of fine-grained actions}

In 2024, the Ball Action Spotting task was further expanded to introduce more granular ball-related events, moving from just two action classes (pass and drive) to a more comprehensive set of $12$ different ball-centric actions: \emph{Pass, Drive, Header, High Pass, Out, Cross, Throw In, Shot, Ball Player Block, Player Successful Tackle, Free Kick,} and \emph{Goal}. 
This extension significantly increases the task’s complexity, providing researchers with an opportunity to develop models that can handle a wider range of interactions between players and the ball while maintaining precise temporal localization.

This expansion to $12$ classes in \SNBASYear{2024} reflects a growing emphasis on ball interactions throughout a match, with each action class capturing specific, frequent events that are useful for understanding the flow of the game. The task highlights the diversity of ball interactions and underscores the necessity of precise temporal localization. 
Additionally, the 2024 version of the SoccerNet challenge introduces new difficulties, such as distinguishing between similar events like passes, high passes, or crosses. This requires a fine-grained understanding of both the visual and contextual aspects of the game, setting higher standards for model performance, and pushing for advancements in temporal detection and sequence modeling within sports video analysis.

The \SNBASYear{2024} dataset is composed of the same $7$ games used in the 2023 version, and the same $11{,}041$ annotated timestamps, but now extended with the $12$ classes instead of $2$. 
Similarly, the challenge set comprises $2$ extra games that are withheld for benchmarking and challenge purposes.

\subsection{Summary} 
SoccerNet has evolved into the largest and most comprehensive dataset for action spotting, enabling the sports video analysis community to develop and benchmark action spotting models. Over multiple iterations, it has expanded from basic action localization to fine-grained ball action spotting, driving innovation in precise temporal localization and contextual understanding of actions in association football videos. The continued growth of the SoccerNet challenges ensures that research in this field has continuously advanced each year, addressing increasingly complex aspects of sports video analysis.

\section{Action Spotting Methods}
\label{sec:methods}

\begin{figure}[t]
    \centering
    \includegraphics[width=\linewidth]{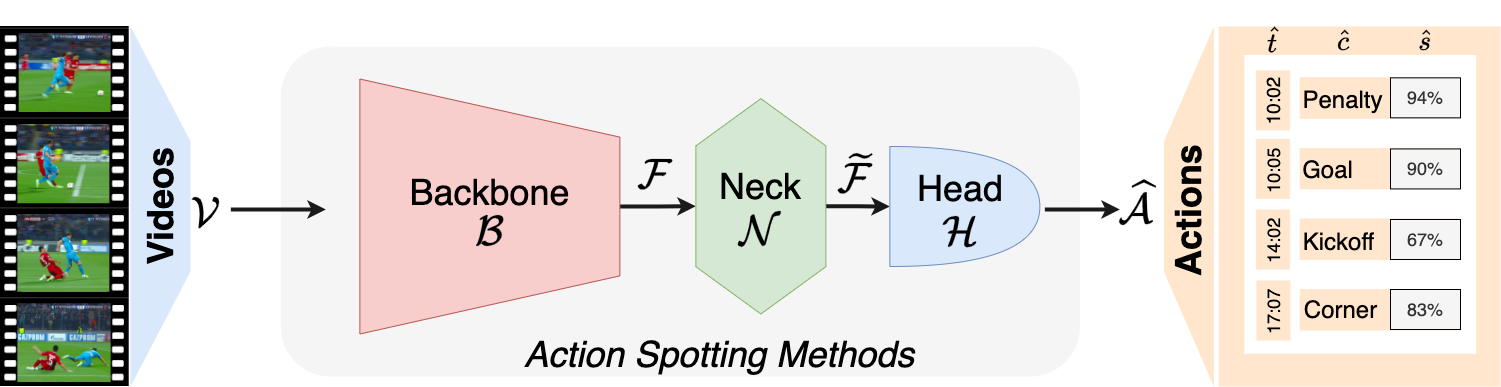}
    \caption{
    \textbf{Generic action spotting method pipeline.}
    A backbone $\backbone$ processes the videos $\videoset$ to extract features $\featureset{}$, refined by the neck $\neck$ into action spotting discriminating features $\neckfeatureset{}$ before being fed into an action spotting head $\head$ that produces action spotting predictions $\predicted{\actionset{}}$, each composed of a timestamp $\predicted{\timestamp}$, a class $\predicted{\class}$, and a confidence score $\predicted{\confidence}$.}
    \label{fig:typicalactionspottingmethods}
\end{figure}

Formally, an action spotting method $\method$ is defined as a function that maps a video $\video{\videoidx}$ to a set $\predicted{\actionset{\videoidx}}=\{\predicted{\action{1}{\videoidx}},\predicted{\action{2}{\videoidx}},...,\predicted{\action{\actionidx}{\videoidx}},...,
\predicted{\action{\predicted{\nactions{\videoidx}}}{\videoidx}}\}$ of $\predicted{\nactions{\videoidx}}$ predicted actions, following $\method: \video{\videoidx} \rightarrow \predicted{\actionset{\videoidx}}$. 
Each predicted action $\predicted{\action{\actionidx}{\videoidx}}$ is a triplet composed of a predicted action class $\predicted{\class}\in\classset$, a timestamp $\predicted{\timestamp}\in[0,\videolength]$, alongside an optional confidence score $\predicted{\confidence}\in[0,1]$ with a default value of $1$. Due to the discrete nature of videos, some methods will predict the timestamp in terms of frame number rather than milliseconds, following $\predicted{\imageidx}\in\{1,2,...,\nimages{\videoidx}\}$. Nevertheless, for evaluation purposes, the prediction is transformed back in millisecond following: $\predicted{\timestamp} = (\predicted{\imageidx}-1) \times \interframe $, with $\interframe$ being the time between two frames.

Typical action spotting methods consist of three primary components: the backbone $\backbone$, the neck $\neck$, and the head $\head$, as illustrated in Figure~\ref{fig:typicalactionspottingmethods}. We detail the role of each component hereafter.
The \emph{backbone} serves as the core component for feature extraction. Given a video $\video{\videoidx}$ composed of frames $\{\image{1}{\videoidx},\image{2}{\videoidx}, ..., \image{\imageidx}{\videoidx},..., \image{\nimages{\videoidx}}{\videoidx}\}$, the backbone processes each frame and extracts a set of features $\featureset{\videoidx} = \{\feature{1}{\videoidx},\feature{2}{\videoidx}, ..., \feature{\imageidx}{\videoidx},..., \feature{\nimages{\videoidx}}{\videoidx}\}$, following $\backbone: \video{\videoidx} \rightarrow \featureset{\videoidx}$. These features form the foundation of the action spotting representation, capturing the essential visual information from the video.
Following the backbone, the \emph{neck} acts as an intermediary layer that refines and transforms the raw features $\featureset{\videoidx}$ into a more discriminative form $\neckfeatureset{\videoidx}$ for action spotting and adapts the feature sizes to accommodate the input format of the head following $\neck: \featureset{\videoidx} \rightarrow \neckfeatureset{\videoidx}$. 
These refined features $\neckfeatureset{\videoidx}$ are then passed to the head.
The \emph{head} is the final stage of the algorithm, responsible for action identification and temporal localization. It uses the refined features $\neckfeatureset{\videoidx}$ to identify and localize the actions $\predicted{\actionset{\videoidx}}$ following $\head: \neckfeatureset{\videoidx} \rightarrow \predicted{\actionset{\videoidx}}$.
Together, these components form a complete action spotting method $\method= \head \circ \neck \circ \backbone$, where $(g \circ f)(x)$ denotes the composition $g(f(x))$.
Some action spotting methods later apply some post-processing on the action set, such as Non-Maxima Suppression (NMS) to remove redundant predictions.

The action spotting methods developed in the last few years can be broadly categorized into \emph{feature-based} and \emph{end-to-end} approaches. In \emph{feature-based} methods, the backbone $\backbone$ consists of pre-trained features extractor from commonly used classification networks like ResNet~\cite{He2015DeepResidual}, trained on large datasets such as ImageNet~\cite{Deng2009ImageNet}. In this case, only the neck $\neck$ and head $\head$ components are trained for the action spotting task. This approach is computationally efficient, but the action spotting performance is highly dependent on the quality of the pre-extracted features for discriminating actions.
In contrast, \emph{end-to-end methods} train the entire architecture, including the backbone $\backbone$, neck $\neck$, and head $\head$, directly on the action spotting task. These methods often provide better performance since the backbone features are optimized for spotting actions on sports videos. However, they generally require significantly larger amounts of data and computational resources to be trained compared to feature-based methods.

In the following, we first describe several historical \emph{feature-based methods} that utilize pre-trained backbones such as ResNet~\cite{He2015DeepResidual} to extract visual features before training task-specific neck and head components for action spotting. Following this, we explore more recent \emph{end-to-end methods}, which jointly aim to optimize the backbone, neck, and head components. Through this progression, we illustrate the evolution of methods proposed for action spotting.

\subsection{Feature-based methods}
\label{subsec:feature-basedMethods}

\mysection{NetVLAD: learnable pooling layers~\cite{Giancola2018SoccerNet}.}

The NetVLAD method was introduced as the baseline for the action spotting task in SoccerNet-v1~\cite{Giancola2018SoccerNet}, leveraging the popular NetVLAD~\cite{Arandjelovic2018NetVLAD} layer originally designed for image retrieval. NetVLAD is particularly effective for summarizing large amounts of visual data into compact representations, which made it a suitable choice for the \SN{1} dataset, where the task was to identify key moments such as goals, cards, and substitutions within long, untrimmed video sequences.
The NetVLAD approach can be described as a feature-based method, where the backbone is pre-trained on large image datasets, and the focus is placed on summarizing video frames into discriminative representations for action spotting.

In NetVLAD, the backbone consists of pre-extracted visual features from a ResNet model~\cite{He2015DeepResidual}, which was trained on the ImageNet dataset~\cite{Deng2009ImageNet}. Instead of training the backbone on football videos directly, the features are extracted frame-by-frame from the input video. This approach significantly reduces computational overhead, as it avoids the need for feature extraction during training and testing.
The features are eventually reduced in dimension through Principal Component Analysis (PCA) and then fed into the NetVLAD module, which acts as the neck of the model, aggregating the extracted frame-level features into a fixed-length representation. Instead of focusing on exact temporal localization, NetVLAD aggregates temporal information over a sampled window of video frames. The process involves:
(i) extracting a temporal video clip from the untrimmed video in a sliding window fashion,
(ii) assigning each frame-level feature to a set of learnable cluster centers, 
(iii) computing the residual between each feature and its assigned center,
(iv) summing up these residuals to produce a compact, video clip-level representation.
This process condenses the temporal information from a candidate video clip into a fixed-dimensional vector, which aggregates the temporally local receptive field.
The head focuses solely on action classification. Once the feature representation pooled from a video clip is obtained, the head applies a multi-class classification layer to determine whether any of the predefined action classes (\eg, goal, card, substitution) occurs within the clip.
Since the model samples video clips from the untrimmed video in a sliding window fashion, it classifies each window without directly predicting the exact timestamp of the actions. Instead, the method outputs whether a particular action occurs within the sampled window.
To manage the untrimmed and lengthy nature of football videos, NetVLAD applies a sliding window approach, producing an \emph{actionness score} in time, \ie, how likely the window represents an action of a certain class. In practice, the video is divided into overlapping temporal windows, and the model classifies each window independently. After classification, Non-Maxima Suppression (NMS) is used over the actionness score in time to refine the predictions by eliminating redundant detections across the overlapping windows. This ensures that only the most confident predictions are retained, reducing false positives and providing a cleaner set of action predictions.

The NetVLAD method served as an important baseline for the first version of SoccerNet. While it does not explicitly perform explicit temporal localization, it provided a practical and efficient framework for action spotting following a classification approach within a sliding window, demonstrating that pre-extracted features combined with temporal pooling could identify key moments in football matches to some extent. Its simplicity and computational efficiency made it a popular choice for early research, though it left room for the development of more advanced methods that could explicitly handle temporal localization and end-to-end training.

%\bigskip

\mysection{NetVLAD++: temporally-aware pooling~\cite{Giancola2021Temporally}.}

NetVLAD++ was introduced as an improvement over the original NetVLAD method for action spotting, addressing a key limitation: the lack of temporal awareness regarding the context surrounding the actions in football broadcasts~\cite{Giancola2021Temporally}. In NetVLAD, all frames in a video clip are treated equally, without distinguishing the context before or after an action. However, for many football events, the preceding and succeeding moments can provide different contextual cues for identifying actions. For instance, spotting a goal from a shot is ambiguous until the succeeding context is revealed. NetVLAD++ introduces a more sophisticated pooling strategy by learning separate clusters for the frames before and after the action, effectively making the model aware of temporal dynamics.

Like its predecessor, NetVLAD++ utilizes pre-extracted frame-level features from a ResNet encoder trained on the ImageNet dataset. The features are extracted from each frame of the video, serving as the raw input to the NetVLAD++ module. This keeps the model computationally efficient by focusing on feature aggregation rather than training a new backbone from scratch.
The primary innovation in NetVLAD++ lies in its temporally-aware pooling mechanism. Rather than pooling features uniformly across the video clip, NetVLAD++ splits the candidate video clip from the sliding window into 2 parts: (i) the frames leading up to the action and (ii) the frames following the action.
For each of these regions, NetVLAD++ learns separate NetVLAD clusters, allowing the model to capture the different types of information conveyed in each temporal segment. This separation helps the model focus on the action’s immediate surroundings, which are often indicative of its occurrence. Following our goal example, the buildup (context before) and aftermath (context after) provide different information that complements the action itself.
The NetVLAD++ layer thus pools features in a temporally-aware manner, creating separate, discriminative representations for the context before and after the action. This improvement allows the model to be more sensitive to the nuances of football events that often rely on contextual buildup or consequences.
Once the temporally-aware features are pooled, the head of the model performs the same action classification along sliding windows with the Non-Maxima Suppression (NMS), refining the action spots. 

The addition of temporally-aware feature pooling in NetVLAD++ significantly improves the performance of action spotting, particularly in cases where the context plays a crucial role in recognizing actions. By learning separate clusters for the pre-action and post-action frames, NetVLAD++ surpasses the performances of the original NetVLAD in detecting actions.
This refinement paved the way for more advanced temporal modeling techniques and remains an influential approach in sports video understanding.

\mysection{CALF: a context-aware loss function for action spotting~\cite{Cioppa2020AContextaware}.}

The CALF (Context-Aware Loss Function) method, developed by Cioppa~\etal~\cite{Cioppa2020AContextaware} introduces a novel approach for action spotting by focusing on the temporal context around actions. The two main contributions are (1) the direct regression of the timestamp value rather than classifying each video chunk followed by a NMS post-processing as previously done in NetVLAD, and (2) the introduction of a context-aware loss function, which weights differently the loss associated to the frames surrounding a ground-truth action during training. The method considers overlapping $2$-minutes video clips rather than ingesting the whole untrimmed video.

For the backbone, CALF uses frozen pre-trained feature extractors, like ResNet, to process each video frame and reduce the input dimensionality. These extractors, originally trained on large datasets such as ImageNet, output a set of features that encapsulate the visual information about each frame. Since the backbone is frozen, the features are extracted once and then stored as matrices with two dimensions: time and number of features. Following NetVLAD, the feature dimension is reduced through Principal Component Analysis (PCA) by a factor of four.
The neck component in CALF is implemented as a trainable temporal convolutional neural network (CNN) that processes the features extracted by the backbone. This temporal CNN consists of multiscale convolutions that operate across time and feature dimensions. The purpose of this component is to fine-tune the raw spatial features by incorporating temporal dependencies across frames.
In CALF, the head is composed of two modules. The first module is a segmentation head that predicts an actionness score for each frame, similar to NetVLAD. The segmentation head is guided during training by the context-aware loss function, which assigns a weight to the cross-entropy loss of each frame depending on the closest ground-truth action location. Specifically, if a frame is located just after an action, CALF assumes that there are still enough visual cues that the action occurred for the segmentation head to predict an action, with a smaller weight the further the frame is from the action. Just before the action, it is unsure whether the action will occur or not, so the weight is set to zero, letting the segmentation module decide freely whether to predict an action. Finally, the further away from the action, the more the segmentation head should predict that no action occurred. This loss therefore alleviates the issue of two similar consecutive frames needing to predict different actions. The predictions of the segmentation module are then concatenated with the neck features and passed to the second module, the spotting head. This head is composed of classical convolutions and max pooling operations to reduce dimensionality and format the output following a YOLO-like~\cite{Redmon2016YOLO} encoding. Specifically, each video clip may predict a finite set of actions with a confidence score, a regressed timestamp representing the relative location of the action within the video clip, and the action class through a softmax activation. The ground-truth and predicted actions are associated through an iterative one-to-one matching algorithm. Successively, a mean squared error loss is operated on the confidence scores and timestamps, and a binary cross entropy on the class.

This method significantly improved upon NetVLAD on the first action spotting task of SoccerNet and the context-aware loss function has been successfully applied on top of other temporal activity localization models such as BSN~\cite{Lin2018BSN} trained on the ActivityNet~\cite{Caba2015ActivityNet} dataset. The context-aware loss also allows for the discovery of actions that are similar to actions of interest, such as goal tentatives thanks to the allowed doubt introduced before the action occurs.

\mysection{Baidu features and transformers~\cite{Zhou2021Feature-arxiv}.}

Zhou~\etal~\cite{Zhou2021Feature-arxiv} propose a novel approach for the two-stage framework of action spotting, integrating fine-tuned backbones with transformer-based temporal detection necks and heads.
Unlike previous methods like NetVLAD++ and CALF, which used generic ResNet features, this method fine-tunes multiple action recognition models specifically for football videos, ensuring improved performance in detecting key moments with greater precision.

The backbone $\backbone$ is a mixture-of-model between the encoders TPN~\cite{Yang2020Temporal}, GTA~\cite{He2021GTA}, VTN~\cite{Neimark2021Video}, irCSN~\cite{Tran2019Video}, and I3D-Slow~\cite{Feichtenhofer2019SlowFast}, fine-tuned on 5-second snippets from the \SN{2} dataset, capturing the $17$ classes of football events.
This fine-tuning tailors the feature extractors to football dynamics, improving performance in detecting key actions. These features, extracted from a stack of frames rather than individual frames, are concatenated to provide richer inputs for temporal detection.
For the neck $\neck$, this method uses a transformer encoder consisting of three encoding layers with sine and cosine positional encodings. Each layer has four heads and a hidden dimension of $64$. The transformer processes the combined features from the backbone, learning temporal dependencies between frames. This allows the model to capture subtle yet important contextual information for action spotting. 
The head $\head$ of the model focuses on action classification by directly predicting the $18$ action class probabilities from the transformer-encoded features ($17$+ background). To refine the output, the method applies hyperparameter tuning, including adjustments to batch size, learning rate, feature dimension size, and the Non-Maxima Suppression (NMS) window size, ensuring that the model selects the most relevant predictions. 

This method achieved the best performance in the first SoccerNet challenge in 2021, based on the \SN{2} dataset, excelling in both action spotting and replay grounding. Its combination of fine-tuned feature extraction and transformer-based temporal modelling raised the standard for action spotting, particularly in handling football’s video manifold distribution. It demonstrated that combining deep attention mechanisms with domain-specific features improves action spotting in sports videos.

\mysection{Dense Detection Anchors: precise action spotting in soccer videos~\cite{Soares2022Temporally}.}

The method presented by Soares~\etal~\cite{Soares2022Temporally} won the SoccerNet challenge 2022, based on the \SN{2} dataset. They introduce a novel approach to action spotting by focusing on precise localization of actions in football videos using dense detection anchors. This method significantly improves the temporal precision of action spotting compared to previous techniques by generating numerous temporal anchors and refining predictions through dense detection.

As with previous methods, Dense Detection Anchors relies on pre-extracted features from a backbone network $\backbone$, either the ResNet~\cite{He2015DeepResidual} trained on ImageNet or Zhou~\etal~\cite{Zhou2021Feature-arxiv} features fine-tuned on SoccerNet. These frame-level features provide the input for the detection process. 
The core innovation of this method lies in the dense temporal anchoring strategy. The video is divided into a set of overlapping temporal segments, and within each segment, the method generates dense temporal anchors at regular intervals. These anchors serve as candidates for potential actions. By densely covering the timeline, the model ensures that no significant action is missed, even in the occurrence of several actions within short durations.
For each anchor, features are extracted from the pre-extracted frame-level representations, and the model evaluates whether an action occurs within that temporal window. The neck component applies temporal convolutions to refine the anchor features and enhance their ability to discriminate actions across time.
Once the anchor features are refined, the head of the model performs two tasks: 
(i) action classification: for each anchor, the model predicts the likelihood of an action occurring. The classification head outputs the predicted class for the action (\eg, goal, card, substitution).
(ii) temporal regression: in addition to classification, the model also refines the temporal localization of the action by regressing the precise timestamp relative to the anchor. This enables the model to predict the exact moment when the action happens, rather than simply classifying a broad temporal window. The dense anchoring and regression mechanisms ensure high temporal precision, improving the performances of action localization.

The method uses a multi-task loss function that combines classification and regression objectives. 
The classification loss penalizes incorrect action predictions while encouraging precise temporal localization, ensuring that the model identifies the correct action and pinpoints the exact moment it occurs, minimizing temporal ambiguity.
After processing all dense detection anchors, Non-Maxima Suppression (NMS) is applied to remove redundant predictions. 

The Dense Detection Anchors method was the top performer in the SoccerNet challenge 2022, significantly improving temporal precision in action spotting. Its high performance in handling complex football actions established it as a state-of-the-art technique for action spotting in sports video analysis.

\subsection{End-to-end methods}
\label{subsec:End-to-endMethods}

\mysection{E2E-Spot: spotting temporally precise, fine-grained
events in video~\cite{Hong2022Spotting}.}

Hong~\etal~\cite{Hong2022Spotting} introduced the first end-to-end model specifically designed for the action spotting task, named \emph{E2E-Spot}. This method brings several novel contributions to the field, particularly in its ability to reason both globally and locally about actions within a video. One of the key innovations of E2E-Spot is its compact architecture, which allows it to be trained on large temporal windows even with limited GPU memory. Despite its compactness, the method demonstrated impressive performance, securing second place in the SoccerNet challenge 2022 on Action Spotting~\cite{Giancola2022SoccerNet}.

Unlike many of the earlier methods discussed in Section~\ref{subsec:feature-basedMethods}, which relied on pre-trained and frozen backbones, the backbone of E2E-Spot is fully trainable in an end-to-end manner. The authors explored several lightweight backbone architectures, including ResNet~\cite{He2015DeepResidual} (18 and 50), RegNet-Y~\cite{Radosavovic2020Designing} (200MF and 800MF), and ConvNeXt~\cite{Liu2022AConvNet}. Another significant improvement in their approach was the integration of temporal information directly within the backbone. This was achieved by incorporating TSM~\cite{Lin2019TSM} (Temporal Shift Module) or GSM~\cite{Sudhakaran2020GateShift} (Gate-Shift Module), which allow the model to capture temporal dynamics across frames. By fine-tuning the backbone specifically for the action spotting task, E2E-Spot gains a significant advantage over methods that rely on backbones trained on unrelated datasets.

The neck of the E2E-Spot model is designed to perform long-term temporal reasoning, providing a global understanding of the video sequence. To achieve this, the authors experimented with several architectures for this module, including GRU~\cite{Chung2014Empirical} (Gated Recurrent Unit), deeper GRU, MS-TCN~\cite{AbuFarha2019MSTCN} (Multi-Scale Temporal Convolutional Network), and ASFormer~\cite{Yi2021ASFormer}. These architectures help the model aggregate information over long sequences, capturing the context necessary to accurately identify and localize actions within the video.

Finally, the head of E2E-Spot is simply a dense frame prediction. For every input frame, the model generates a corresponding output feature and a per-frame prediction. During training, a simple cross-entropy loss is applied to each frame's prediction using a one-hot encoding of the ground-truth action class, along with an extra class for background frames. This straightforward approach ensures that the model is directly optimized for frame-level accuracy. The results demonstrate that direct, end-to-end training of a simple and compact model like E2E-Spot can serve as a surprisingly strong end-to-end baseline for action spotting.

\mysection{Gray-scale image stacking and transfer learning to cope with limited data~\cite{Baikulov2023Solution}.}

Ruslan Baikulov's method, which emerged as the winner of the \SNBASYear{2023}  challenge~\cite{Baikulov2023Solution}, showcases several innovative approaches tailored to the specific challenges of detecting ball-related actions. One of the key features of this method is the use of stacked grayscale images. In this approach, three consecutive frames are first converted from RGB to grayscale and then concatenated to form a new three-channel image. This transformation allows the model to focus on motion cues rather than color information, which is particularly advantageous for detecting subtle ball-related events where motion is a critical indicator.

The backbone of Baikulov's method employs a 2D convolutional encoder, EfficientNetV2-B0~\cite{Tan2021EfficientNetV2}, to extract per-frame features. These features are then passed to the neck, which uses a 3D convolutional encoder to merge temporal information across consecutive frames, effectively capturing motion dynamics. The merged features are concatenated along the temporal dimension and pooled using Generalized Mean (GeM) pooling to condense the information. Finally, the head consists of a simple linear classifier that predicts the presence and types of ball-related actions based on the refined features of the neck.

Given the limited amount of data available in the Ball Action Spotting challenge, Baikulov's method leverages transfer learning to maximize performance. The 2D convolutional encoder is initialized with pre-trained weights from ImageNet~\cite{Deng2009ImageNet}, while the 3D convolutional encoder and the linear classifier are trained from scratch. Initially, the model is trained on short sequences of $15$ frames to manage computational complexity and memory constraints. The model is first pre-trained on the larger and more diverse \SN{2} dataset, which contains more videos and annotations, though with different action classes. After this initial training, transfer learning is applied by fine-tuning the model on the specific \SNBASYear{2023} dataset. Finally, to further enhance performance, the model undergoes fine-tuning on longer sequences of 33 frames, with the 2D backbone frozen to prevent overfitting and to allow the model to focus on longer temporal dynamics.

\mysection{T-DEED: temporal-discriminability enhancer encoder-decoder for precise event spotting in sports videos~\cite{Xarles2024TDEED-arxiv}.}

T-DEED, the winning method of the \SNBASYear{2024} challenge~\cite{Cioppa2024SoccerNet2024-arxiv}, introduced by Artur Xarles~\etal~\cite{Xarles2024TDEED-arxiv}, tackles several key challenges in action spotting, including improving the discriminability of frame representations, achieving high temporal resolution, and capturing information across multiple temporal scales to handle diverse action dynamics. This method represents a significant architectural advancement over previous approaches.

The backbone of T-DEED is designed to process fixed-length video clips and is built on a 2D RegNetY~\cite{Radosavovic2020Designing} architecture, augmented with Gate-Shift-Fuse~\cite{Sudhakaran2023GateShiftFuse} (GSF) modules. This backbone therefore produces spatio-temporal per-frame representations. 
The neck component of T-DEED consists of a temporally-discriminant encoder-decoder architecture, which leverages Scalable-Granularity Perception~\cite{Shi2023TriDet} (SGP) layers. The authors also include skip connections to preserve information from the initial layers. Additionally, the authors introduce a novel SGP-Mixer layer that accommodates inputs with distinct temporal scales.
Finally, the refined features from the neck are passed to a classical action spotting classification head~\cite{Hong2022Spotting,Soares2022Temporally,Xarles2023ASTRA}. This head includes both a classification and displacement head, responsible for predicting the presence of actions and their precise temporal locations within the clip. 

The combined architecture of T-DEED, with its focus on multiscale temporal processing and enhanced frame-level discriminability, marks a significant improvement over previous architectures in the domain of precise action spotting.

\section{Evaluation and Metrics}
\label{sec:metric}
An action spotting method $\method$ predicts a set of actions $\predicted{\actionset{\videoidx}}$ for each video $\videoidx$. To evaluate the performance of $\method$, the set of predicted actions needs to be compared to the set of ground-truth actions for that video $\actionset{\videoidx}$. An oracle method would therefore satisfy the following condition: $\predicted{\actionset{\videoidx}} = \actionset{\videoidx}, \forall \videoidx \in \{1,...,\nvideos\}$. Giancola~\etal~\cite{Giancola2018SoccerNet} proposed a first metric, called the average-mean Average Precision ($\amap$) which is inspired by classical object detection metrics in PASCAL VOC~\cite{Everingham2010PascalVOC} as well as temporal activity localization metrics from the ActivityNet~\cite{Caba2015ActivityNet} challenges. 
This metric and some of its variants are still used nowadays to evaluate action spotting methods, including the SoccerNet challenges. The remaining of this section details the procedure to compute the $\amap$ and its variants.

The first step consists in associating elements from the set of predicted actions $\predicted{\actionset{\videoidx}}$ one-to-one to the set of ground-truth actions $\actionset{\videoidx}$ for each video $\videoidx$ of the evaluation dataset. 
To do so, Giancola~\etal~\cite{Giancola2018SoccerNet} proposed an iterative one-to-one matching algorithm. 
Since the predicted action set contains confidence scores $\predicted{\confidence}$ associated with each action, they should be considered in the evaluation procedure. 
The $\amap$ evaluates a parametric family of action spotting models, depending on a chosen confidence score threshold $\confidencethreshold$, which is a similar paradigm to what is typically done in the evaluation of object detection models.
Therefore, one matching per considered threshold $\confidencethreshold$ is first performed. 
Following the practice of Everingham~\etal~\cite{Everingham2010PascalVOC}, $200$ threshold values equally spaced between $[0,1[$ are chosen.
For each threshold $\confidencethreshold$, a single ground-truth action $\action{\actionidx}{\videoidx}$ may be matched with one predicted action $\predicted{\action{\actionidx'}{\videoidx}}$ only if three conditions are met: (1) the confidence score of that prediction is greater or equal than the threshold, \ie, $\predicted{\confidence} > \confidencethreshold$, (2) they belong to the same class, \ie, $\predicted{\class}=\class$, and (3) the prediction falls within a $\timeinterval$ time tolerance around the ground-truth action, \ie, $|\predicted{\timestamp} - \timestamp| \le 0.5\times\timeinterval$. If the ground-truth action has several associated predictions, only the prediction with the highest confidence is matched with that ground-truth action.
The pair of matched $(\action{\actionidx}{\videoidx}, \predicted{\action{\actionidx'}{\videoidx}})$ is then removed from the pool of available predicted/ground-truth actions and placed in a set of true-positive pairs $\truepositiveset{\class}{\delta}{\confidencethreshold}$. If no predictions respect the conditions, the ground-truth action $\action{\actionidx}{\videoidx}$ is placed in a set of false negative ground-truth action $\falsenegativeset{\class}{\delta}{\confidencethreshold}$ The matching algorithm is then performed on the next ground-truth action until all ground-truth actions are either placed in the $\truepositiveset{\class}{\delta}{\confidencethreshold}$ or the $\falsenegativeset{\class}{\delta}{\confidencethreshold}$ sets. All remaining unmatched predictions $\predicted{\action{\actionidx'}{\videoidx}}$ are placed in a set of false positive predicted actions $\falsepositiveset{\class}{\delta}{\confidencethreshold}$.

For each class $\class$ and time interval $\timeinterval$, a set of Precision-Recall points $\precisionrecall{\class}{\timeinterval}$ is constructed by computing the precision and recall value for each $\confidencethreshold$ following:
%Step by Step process to estimate the metrics (Avg-mAP, tight, mAP@1, etc...)
$$\precisionrecall{\class}{\timeinterval} = \bigl\{(\frac{\ntruepositives{\class}{\timeinterval}{\confidencethreshold}}{{\ntruepositives{\class}{\timeinterval}{\confidencethreshold}}+{\nfalsepositives{\class}{\timeinterval}{\confidencethreshold}}}, 
\frac{\ntruepositives{\class}{\timeinterval}{\confidencethreshold}}{{\ntruepositives{\class}{\timeinterval}{\confidencethreshold}}+{\nfalsenegatives{\class}{\timeinterval}{\confidencethreshold}}}) 
| \confidencethreshold \in \{0,0.005,0.01,...,0.995\}\bigr\}\, .$$
The set of precision-recall points are then summarized in the Average-Precision ($\ap{}{}$) metric following the 11-point approximation proposed in the PASCAL VOC challenge~\cite{Everingham2010PascalVOC}. 
Specifically, this metric computes an approximation of the area under the precision-recall curve following:
$$\ap{\class}{\timeinterval} = \frac{1}{11} \sum_{r'=0}^{10} \max_{ (\precision, \recall) \in \prsubset(r')} p \quad \text{with} \quad \prsubset(r') =  \bigl\{(\precision, \recall) \in \precisionrecall{\class}{\timeinterval} |  \recall \ge \frac{r'}{10}\bigr\}\, .$$
Afterward, the $\ap{\class}{\timeinterval}$ are averaged across all classes into the mean-Average Precision ($\map{\timeinterval}$) metric following:
$$\map{\timeinterval} = \frac{1}{\nclasses} \sum_{\class=1}^{\nclasses} \ap{\class}{\timeinterval}\, .$$
This $\map{\timeinterval}$ provides an evaluation of how an action spotting method identifies and localizes actions given a specific time tolerance $\timeinterval$ around the ground-truth actions. The smaller the time tolerance, the more precise an action spotting method needs to be.
This metric has proven particularly useful in the ball action spotting benchmarks and for ranking methods in the challenges, as actions related to the ball are very localized in time and therefore need to be precisely predicted. Hence, the $\map{1}$ is used to rank ball action spotting methods, meaning that actions need to be localized around the ground-truth actions within a $1$-second tolerance.

Finally, Giancola~\etal~\cite{Giancola2018SoccerNet} propose to summarize the $\map{\timeinterval}$ over a set of $\timeinterval$ values into a single metric by computing an approximation of the area under the $(\timeinterval,\map{\timeinterval})$ curve. To do so, they rely on a trapezoid integral approximation considering $\ntimeintervals$ time intervals equally spaced between $\timeintervalstart$ and $\timeintervalstop$ time tolerances. They define this new metric as the Average-mean Average Precision ($\amap$). The general formula to compute the $\amap$ is given by:

$$\amap = \frac{1}{\ntimeintervals} \sum_{d=1}^{\ntimeintervals} \frac{\map{(\timeintervalstart + (d-1)\times \frac{\timeintervalstop-\timeintervalstart}{\ntimeintervals})} + \map{(\timeintervalstart + d \times \frac{\timeintervalstop-\timeintervalstart}{\ntimeintervals})}}{2}\, . $$

The first metric introduced by Giancola~\etal~\cite{Giancola2018SoccerNet}, later called the loose Average-mean Average Precision ($\amap_{loose}$), considered $\ntimeintervals = 11$ time intervals of $5$ seconds between tight time tolerances ($\timeintervalstart = 5$ seconds) to loose ones ($\timeintervalstop = 60$ seconds). 
A second stricter variant was later proposed during the \SNAS 2022 and 2023 challenges~\cite{Cioppa2024SoccerNet2023Challenge} considering tighter time tolerances with $\ntimeintervals = 4$ time intervals of $1$ second linearly spaced between $\timeintervalstart = 1$ and $\timeintervalstop = 5$ second time tolerances, called the $\amap_{tight}$.
Let us note that for counting applications, the $\amap_{loose}$ or the $\map{\infty}$ are sufficient as predicting the moment when the actions occur is not useful, while for video analysis of specific actions, the $\amap_{tight}$ or the $\map{1}$ would be preferred, depending on the time tolerance required for the downstream application.
As a final remark, SoccerNet-v2~\cite{Deliege2021SoccerNetv2} introduced the concept of visible versus unshown actions, which correspond to actions that are respectively clearly visible on the broadcast or that happened but were not shown, \eg, due to a replay or showing a different angle. The metrics described above can therefore be defined for each set separately, for analysis purposes. This is interesting to analyze whether methods can understand the rules of the game and still detect unshown actions based on context alone.

\section{Benchmark Across Four Datasets: Summarizing 6 Years of Research}
\label{sec:benchmark}

\begin{table}[t]
\tbl{\SN{1}~\cite{Giancola2018SoccerNet} leaderboard for 3 classes: goal, card, and substitution. The methods are ranked according to the test set $\amap_{loose}$ metric. The $\aap_{loose}$ for each class is also displayed when available. \label{tab:LeaderboardSpotting_3}} 
{\begin{tabular}{@{}lcccc@{}} \toprule
& \multicolumn{4}{c}{Test Set}  \\ 
Methods & \multicolumn{1}{c}{$\amap_{loose}$} & \multicolumn{3}{c}{$\aap_{loose}$} \\ %\cline{2-7}
&  & Goal & Card & Substitution \\ \colrule
Nakazawa~\etal~\cite{Minoura2021Action} & \textbf{81.6} & \textbf{87.1} &  \textbf{63.3} & \textbf{94.3} \\
RMSNet~\cite{Tomei2021RMSNet}$^{\text a}$ & 75.1 & / & / & / \\
Si~\etal~\cite{Shi2022Action} & 66.8 & / & / & / \\
Karimi~\etal~\cite{Karimi2022Soccer} & 64.9 & / & / & / \\
Mahaseni~\etal~\cite{Mahaseni2021Spotting} & 63.3 & / & / & / \\
CALF~\cite{Giancola2018SoccerNet}$^{\text b}$  & 62.5 & / & / & / \\
Vats~\etal~\cite{Vats2020Event} & 60.1 & / & / & / \\
Rongved~\etal~\cite{Rongved2021Automated} & 56.3 & 75.1 & 40.1 & 50.21 \\
AudioVid~\cite{Vanderplaetse2020Improved} & 56.0 & / & / & / \\
NetVLAD~\cite{Giancola2018SoccerNet}$^{\text c}$  & 49.7 & / &  / & / \\
Rongved~\etal~\cite{Rongveld2021Using3D} & 32.0 & / & / & / \\
\botrule
\end{tabular}}
{
\begin{tabnote}
$^{\text a}$\RMSNetURL \\
$^{\text b}$\CALFURL \\
$^{\text c}$\NetVLADURL \\
\end{tabnote}
}
\end{table}
Since the creation of SoccerNet in 2018, numerous methods have been proposed to tackle the action spotting task, reflecting the evolution of both the dataset and the technological landscape of artificial intelligence. Initially, most methods were feature-based, leveraging the pre-extracted features provided by the \SN{1} dataset~\cite{Giancola2018SoccerNet}, which focused on three classes of actions. At that time, end-to-end training on video data was challenging due to limitations in GPU VRAM memory and processing speed. As a result, the early methods focused on refining the neck and head components of the pipeline while relying on pre-extracted features from pre-trained backbones like ResNet. The original NetVLAD~\cite{Giancola2018SoccerNet} method set a baseline, but there was a significant gap until the CALF~\cite{Cioppa2020AContextaware} method introduced a context-aware loss function that substantially improved performance. Although several methods attempted to close this gap, it was not until newer architectures, such as attention mechanisms~\cite{Minoura2021Action}, multi-scene encoders~\cite{Shi2022Action}, and metric learning~\cite{Karimi2022Soccer}, were introduced a year later that better performances were achieved. \Tref{tab:LeaderboardSpotting_3} presents the results of these methods on the original action spotting (3 classes) benchmark.

\begin{table}[t]
\tbl{\SN{2} leaderboard for 17 actions. The methods are ranked according to the challenge set $\amap_{tight}$. \textit{Main} considers all actions, \textit{visible} only visible actions, and \textit{unshown} unshown actions in the broadcast.\label{tab:LeaderboardSpotting_17}}
{\resizebox{\columnwidth}{!}{%
\begin{tabular}{lccc|ccc|ccc|ccc}  \toprule
% Title
& \multicolumn{6}{c|}{Test Set} & \multicolumn{6}{c}{Challenge Set}  \\ \midrule
Methods & \multicolumn{3}{c}{$\amap_{tight}$} & \multicolumn{3}{c}{$\amap_{loose}$} & \multicolumn{3}{c}{$\amap_{tight}$} & \multicolumn{3}{c}{$\amap_{loose}$}\\ %\cline{2-7}
& main & visible & unshown & main & visible & unshown & \textbf{main} & visible & unshown & main & visible & unshown \\ \colrule
% Teams
\bf{MEDet}  & / & / & / & / & / & / & \bf{71.31} &  76.29  & 54.09 & 78.56  & 81.67 & 69.13\\

mt\_player  & / & / & / & / & / & / & 71.10 & \textbf{77.22}  & 58.50 & 78.79  & \textbf{82.02} & 77.62 \\

ASTRA~\cite{Xarles2023ASTRA}$^{\text a}$  & / & / & / & /& / &/  &  70.10 & 75.00  & 57.98 & \textbf{79.21}  & 81.69 &  75.36 \\

HCMUS-PK~\cite{Tran2024Unifying-arxiv}$^{\text b}$  & 62.49 & 69.04 & 	30.42 & 73.98 & 	80.09 & 45.05 &  69.38 & 75.50  & 	53.31 & 76.15  & 80.08 & 66.62 \\

team\_ws\_action  & / & / & / & / & / & /  & 69.17 & 75.18 & 59.12 & 76.95 & 80.39 & 75.92\\

COMEDIAN~\cite{Denize2024COMEDIAN}$^{\text c}$  & 73.1 & / & / & / & / & / &68.38 &74.79 &	47.68 &	73.98 & 78.57 &	61.75\\

Soares~\etal~\cite{Soares2022Temporally}$^{\text d}$  & 	65.07 & 	71.24 & 	36.62 & 78.59 & 83.89 & 	54.15 & 68.33 & 73.22  & \textbf{60.88} & 78.06  & 80.58 & \textbf{78.32} \\

DVP   & / & / & / & / & / & / &66.95 &	74.68 &	53.81 &	73.61 &	79.15 &	67.38 \\

E2E-Spot~\cite{Hong2022Spotting}$^{\text e}$ & / & / & /& / & / &/  &66.73 &	74.84 &	53.21 &	73.62 &	79.16 &	67.42 \\

AS\&RG & / & / & / & / & / & / & 64.88 &	70.31  &	53.03 &	72.83 &	76.08 &	72.35 \\

MSAction & / &/ &/  & / & / & / & 62.26 &	67.48  & 45.04 & 69.86 &	73.81 &		59.15 \\

Faster-TAD~\cite{Chen2022FasterTAD-arxiv} & 61.10 & 25.50 & 54.09 & / & / & / & / & / & / & / & / & / \\

Rkrystal  & 60.94 & 67.49 & 30.97 & 76.07 & 81.54 &  48.64 & 61.84 & 67.39  & 48.71 & 74.75 &	78.29 &	69.02 \\

arturxe   & 	57.28 & 	63.57 & 29.04 & 72.13 & 77.11 & 	45.62 & 60.56 & 65.75  & 53.00 & 71.72 &	75.15 &69.91 \\

cihe   & 	60.51 & 	66.17 & 31.91 & 75.33 & 80.20 & 46.74 & 59.97 & 	64.51  & 53.80 & 72.95 &	76.29 & 71.95 \\

SpotFormer~\cite{Cao2022SpotFormer} & 60.9 & 48.6 & 31.0 & 81.5 & 48.6 & 31.9 &/&/&/&/ & / & / \\

STE-v2~\cite{Darwish2022STE}    & 58.29 & 	63.13 & 30.60 & 71.58 & 	75.36 & 	46.99 & 58.71 & 	63.70  & 51.86 & 70.49 &	73.46 & 	70.11 \\

intro- and inter    & / & / & / & / & / & / & 53.30 & 	58.48  & 47.30 & 66.59 & 68.75 & 67.09 \\

memory     & / & / & / & / & / & /  & 52.89 & 	57.82  & 49.85 & 68.00 & 70.58 & 69.20 \\

Zhu~\etal~\cite{Zhu2022ATransformerbased}$^{\text f}$     & / & / &  / & / & / & / & 52.04 & 	60.18  & 32.06 & 60.86 & 66.64 & 48.46 \\

JAMY2 (AF\_GRU)  & / &/ &/ &/ &/ &/  &	51.97 	&58.05 	&44.29 	&63.12 	&65.98 &	61.66 \\

tyru (GRU\_CALF)  & /& /& /& /& /& / &	51.38 	&57.50 	&41.82 	&62.88 	&66.30 &	56.57 \\

Zhou~\etal~\cite{Zhou2021Feature-arxiv}$^{\text g}$   & 47.05 & 53.33 & 25.63 & 73.77 & 79.28 &  47.84 &	49.56 	&54.42 	&45.42 	&74.84 	&78.58 &	71.52 \\

JAMY (LocPoint)  & /& /& /& /& /& / &	45.83 	&49.68 	&45.71 	&61.80 	&64.23 &	63.48 \\

zqing   & 53.63 & 	56.11 & 44.47 & 	76.08 & 78.10 & 67.99 &	45.76 	&50.37 	&	45.41 	&66.13 	&68.09 &	70.06 \\

SIT    & /& /& /& /& /& / &	21.60 	&26.55 	&	16.83 	&29.92 	&34.92 &	25.22 \\

test\_YYQ   & /& /& /& /& /& / &	12.73 &	14.13 &	11.21 	&54.21 	&58.75 	&48.55 \\ 

ABCS    & /& /& /& /& /& / &	12.22 &	13.55 &	13.80 	&57.16 	&62.25 	&	50.25 \\ 

Visual Analysis of Humans   & / &/  &/  &/  &/  &/   &/	 &	/ &	/ 	&64.73 	&	67.96 	&	50.14 \\ 

RMS-Net~\cite{Tomei2021RMSNet}$^{\text h}$   &  /& / & / & 63.49 & 68.88 &  38.02 & /	 &	/ &	/ 	&60.92 	&	64.09 	&	56.61 \\

MA-VLAD~\cite{Feng2024MAVLAD}  & / & / & / & 62.5 & 67.1 & 39.6 & /& / & / &/ &/ &/ \\

Shi~\etal~\cite{Shi2022Action} & / & / & / & 55.2 & / & /& / &/ &/ &/ &/ & / \\

IdealCat   & / & / &/ & /& /& / &	/ &/	 &	/ 	&	54.24 	&57.50 	&	56.54 \\ 

NetVLAD++~\cite{Giancola2021Temporally}$^{\text i}$   & /& /&/ & 53.40 & 59.41 &  34.97 &	 &	/ &	 /	&52.54 	&57.12 	&	46.15 \\ 

straw    & / &/  &/ & 49.79 & 56.35 & 31.14 &	/ &	/  &	 /	&51.65 	&	57.03 	&	45.33 \\ 

CALF+PlayerLoc~\cite{Cioppa2021Camera}$^{\text j}$  & / & / & / & 46.8 & / & / &	/ &	/ &	 /	& /	& /	&	/ \\ 

CALF~\cite{Cioppa2020AContextaware}$^{\text k}$  & / & / & / & 41.61 & 43.54 & 28.88 &	 /&	/ &/	 	&42.22 	&43.51 	&		37.91 \\ 

AudioVid~\cite{Vanderplaetse2020Improved} & / & / & / & 39.9 &  / & /  &	/ &/	 &	/ 	& /	& /	& /\\ 

NetVLAD~\cite{Giancola2018SoccerNet}$^{\text l}$  & / & / & / & 31.37 & 34.30 &  23.27 &	/ &	/ &	 /	&30.74 	&32.99 	&	23.27 \\ 
\botrule
\end{tabular}}%
}
{\begin{tabnote}
$^{\text a}$\ASTRAURL \\
$^{\text b}$\HCMUSPKURL \\
%\footnote{\url{https://github.com/Fsoft-AIC/UGLF}} 
$^{\text c}$\COMEDIANURL \\
%~\footnote{\url{https://github.com/juliendenize/eztorch}}
$^{\text d}$\SoaresURL \\
%~\footnote{\url{https://github.com/yahoo/spivak}} \\
$^{\text e}$\ETWOESpotURL \\
%~\footnote{\url{https://github.com/jhong93/spot}} \\
$^{\text f}$\ZhuURL \\
%~\footnote{\url{https://github.com/ArthurUnic/action_spotting}} \\
$^{\text g}$\ZhouURL \\
%~\footnote{\url{https://github.com/baidu-research/vidpress-sport}} \\
$^{\text h}$\RMSNetURL \\
%~\footnote{\url{https://github.com/aimagelab/RMSNet_Soccer}}
$^{\text i}$\NetVLADPPURL \\
%~\footnote{\url{https://github.com/SoccerNet/sn-spotting/tree/main/Benchmarks/TemporallyAwarePooling}}
$^{\text j}$\CALFPlayerLocURL \\
%~\footnote{\url{https://github.com/SoccerNet/sn-spotting/tree/main/Benchmarks/CALF_Calibration_GCN}}
$^{\text k}$\CALFURL \\
$^{\text l}$\NetVLADURL
\end{tabnote}
}
\end{table}

The release of \SN{2}~\cite{Deliege2021SoccerNetv2}, which  expanded the task to 17 classes, marked a significant step forward and laid the foundation for the first SoccerNet challenge on Action Spotting in 2021. This task provided the research community with a more challenging setup, leading to the development of new techniques. The first major performance boost came from Zhou~\etal~\cite{Zhou2021Feature-arxiv}, who fine-tuned pre-trained features on football data through a frame classification task. In the following year’s challenge~\cite{Giancola2022SoccerNet}, the evaluation metric was tightened from $\amap_{loose}$ to $\amap_{tight}$, requiring methods to focus on precise temporal localization. This led to significant breakthroughs, including the first end-to-end method proposed by Hong~\cite{Hong2022Spotting}. However, a feature-based method by Soares~\etal~\cite{Soares2022Temporally} still emerged victorious, utilizing dense detection anchors to refine spotting predictions. The 2023 edition of the SoccerNet challenges~\cite{Cioppa2024SoccerNet2023Challenge} saw several methods surpass the 70\% mark in $\amap_{tight}$, with the winning entry employing a hybrid architecture that fused transformer and CNN-based encoders. The performances of these methods are summarized in \tref{tab:LeaderboardSpotting_17}.

\begin{table}[t]
\caption{Ball Action Spotting (2023) leaderboard for 2 classes. The methods are ranked according to the challenge set $\map{1}$.\label{tab:LeaderboardBallSpotting_2}}
{\resizebox{\columnwidth}{!}{%
\begin{tabular}{lccccc|c|ccccc|c} \toprule
% Title
& \multicolumn{6}{c|}{Test Set} & \multicolumn{6}{c}{Challenge Set}  \\ \colrule
Methods & \multicolumn{5}{c|}{$\map{}$} & \multicolumn{1}{c|}{$\amap_{tight}$} & \multicolumn{5}{c|}{$\map{}$} & \multicolumn{1}{c}{$\amap_{tight}$} \\ %\cline{2-7}
&  1 & 2 & 3 & 4 & 5 & & \textbf{1} & 2 & 3 & 4 & 5 &  \\ \colrule
% Teams
\textbf{Ruslan Baikulov}$^{\text a}$  & 87.04 & 88.09 & 87.85 & 	87.10 &86.64 &  87.47	&\textbf{86.47} 	&\textbf{87.98} 	&\textbf{88.28} 	&\textbf{88.18} 	&\textbf{87.95} 	&\textbf{87.91}\\
Wang~\etal~\cite{Wang2023ABoosted-arxiv}$^{\text b}$  &86.37 & 87.16 & 87.15 & 	85.80 & 	85.48 &  	86.51	&83.39 	&85.19 	&85.81 	&86.00 	&86.19 	&85.45\\
BASIK & /& /& /& /& /&  /	&82.06 	&83.39 	&83.86 	&84.04 	&83.91 	&83.57\\
FC Pixel Nets  & 	83.53 & 	84.89 & 	85.40 &	84.57 & 84.03&  84.66	&81.89 	&83.22 	&83.97 	&83.85 	&84.02 &	83.50\\
play  & /& /& /& /& /&  	/&79.74 	&82.58 	&84.06 	&84.49 	&84.34 	&83.29\\
E2E-Spot~\cite{Hong2022Spotting}$^{\text c}$  & 69.43&74.45 & 76.11 & 75.52 & 	78.05 &  	74.96	&62.72 	&69.24 	&72.57 	&74.29 	&74.80 	&71.21\\
\botrule
\end{tabular}}
}
\begin{tabnote}
$^{\text a}$\BaikulovURL \\
%\footnote{https://github.com/lRomul/ball-action-spotting}
$^{\text b}$\WangURL \\
%~\footnote{https://github.com/ZJLAB-AMMI/E2E-Spot-MBS}
$^{\text c}$\ETWOESpotURL 
\end{tabnote}
\end{table}
In 2023, we introduced the first edition of the \SNBAS challenge~\cite{Cioppa2024SoccerNet2023Challenge}, focusing on two classes: passes and drives. This new task presented three key challenges: the need to detect fast and subtle actions with minimal visual cues, the density of actions leading to difficulties in distinguishing between closely occurring actions, and the limited amount of training data available, which encouraged the use of semi-supervised, self-supervised, and transfer learning techniques. Participants in this challenge explored various strategies, including pre-training on action spotting videos and fine-tuning on ball action data, utilizing stacked sequences of grayscale images in the RGB channels, employing focal loss for label expansion, and model ensembling to combine different network variants. The results of these methods are detailed in \tref{tab:LeaderboardBallSpotting_2}.

\begin{table}[t]
\tbl{ Ball Action Spotting (2024) leaderboard for 12 classes. The methods are ranked according to the challenge set $\map{1}$.\label{tab:LeaderboardBallSpotting_12}}
{\resizebox{\columnwidth}{!}{%
\begin{tabular}{lccccc|c|ccccc|c} \toprule
& \multicolumn{6}{c|}{Test Set} & \multicolumn{6}{c}{Challenge Set}  \\ \colrule
Methods & \multicolumn{5}{c|}{$\map{}$} & \multicolumn{1}{c|}{$\amap_{tight}$} & \multicolumn{5}{c|}{$\map{}$} & \multicolumn{1}{c}{$\amap_{tight}$} \\ %\cline{2-7}
&  1 & 2 & 3 & 4 & 5 & & \textbf{1} & 2 & 3 & 4 & 5 &  \\ \midrule
% Teams
        T-DEED~\cite{Xarles2024TDEED-arxiv}$^{\text a}$ & / &/ &/ & /& /& / & \textbf{73.39}  & 	\textbf{76.82} & \textbf{77.83} & 	\textbf{78.42} & \textbf{78.51} & \textbf{77.25}  \\
        UniBw Munich - VIS & / &/ &/ & /& /& / & 71.35 & 75.05 & 75.87 & 76.22 & 76.27 & 75.24  \\
        FS-TAHAKOM  & / &/ &/ & /& /& / &  67.09  & 70.22 & 71.26 & 71.74 & 71.59 &  70.64 \\
        MobiusLabs & / &/ &/ & /& /& / &  62.53 &  65.46 & 67.01 & 67.15 & 67.27 & 66.13  \\
        AI4Sports  & 	66.33 & 	69.23 & 	70.75 & 71.24 &	71.44 &  70.03 &  62.44 & 65.53 & 66.49 & 	66.92 & 66.98 &  65.91 \\
        sota  & / &/ &/ & /& /& / & 62.44  & 65.68 & 66.69 & 67.48 & 67.35 &   66.19	  \\
        SAIVA & / &/ &/ & /& /& / & 56.74  & 60.23 & 60.98 &	61.86 & 61.75 & 60.58  \\
        Ruslan Baikulov$^{\text b}$ & 59.70 & 62.73 & 	63.74 & 64.47 & 64.87 & 63.31 & 56.15  & 	60.30 & 61.20 & 61.86 & 61.91 &  60.60 \\
        blueblue111 & / &/ &/ & /& /& / & 3.66  & 	4.08 & 4.12 & 4.17 & 	4.01 &  4.05 \\
\botrule
\end{tabular}}%
}
\begin{tabnote}
$^{\text a}$\TDEEDURL \\
%~\footnote{https://github.com/arturxe2/T-DEED}
$^{\text b}$\BaikulovURL
%\footnote{https://github.com/lRomul/ball-action-spotting}
\end{tabnote}
\end{table}

\medskip

In 2024, the complexity of the ball action spotting task~\cite{Cioppa2024SoccerNet2024-arxiv} was increased by expanding the number of classes from 2 to 12. This expansion required more granular categorization of actions, such as differentiating between various types of passes. Participants employed several advanced techniques to tackle this challenge. Similarly to 2023, key strategies included the use of grayscale images to simplify visual processing, the integration of 2D and 3D convolutional networks to capture both spatial and temporal features, and the application of transfer learning from related domains to enhance model training. The winning team, T-DEED~\cite{Xarles2024TDEED-arxiv}, introduced an innovative approach using a 2D backbone with Gate-Shift Fuse (GSF) modules for generating per-frame representations with local spatiotemporal information. Their method also included a temporally-discriminant encoder-decoder that refined per-frame tokens, increasing their discriminability within sequences while maintaining high temporal resolution. To address the issue of limited data, they trained their model on both \SNBASYear{2024} and the original \SN{2} datasets simultaneously, using a multitask training approach with dual prediction heads. The results of these methods are presented in \tref{tab:LeaderboardBallSpotting_12}.

Overall, the challenges organized around SoccerNet have attracted significant attention and fostered the development of innovative methods that have applications beyond sports video analysis. Notably, some winning methods, such as E2E-Spot by Hong~\etal~\cite{Hong2022Spotting}, Dense Detection Anchors by Soares~\etal~\cite{Soares2022Temporally}, and T-DEED by Xarles~\etal~\cite{Xarles2024TDEED-arxiv}, have shown strong performance on other datasets, demonstrating that advancements made for sports can contribute to breakthroughs in more generic video understanding. This underscores the relevance of the SoccerNet dataset, benchmark, and challenges to the broader research community. By providing open-source data and organizing these challenges, SoccerNet has successfully built a large community of researchers dedicated to advancing the field of video understanding.
\section{Conclusion}
\label{sec:conclusion}

In this chapter, we provided a comprehensive overview of the task of \emph{action spotting}, structured around three key pillars: datasets, methods, and evaluation metrics. We began by tracing the evolution of action spotting, highlighting the important role of the SoccerNet datasets in initiating extensive research in this domain. The introduction of these large-scale datasets, completed with detailed annotations, has been crucial in driving forward the field by providing a solid foundation for the development of innovative approaches.

Next, we explored how deep learning methods progressively approached the task of action spotting, discussing the general architecture and then describing specific methods published in the literature. Specifically, we examined both feature-based approaches, which rely on pre-trained backbones, and end-to-end methods that optimize the entire model architecture for the task. These methods have introduced significant advancements, each contributing unique insights and techniques that have pushed the state of the art in action spotting.

We also detailed the evaluation metrics, which are essential for assessing the performance of action spotting methods, emphasizing the importance of precise temporal localization and the challenges it presents. Following this, we presented a retrospective analysis of six years of research, showcasing a thorough benchmark of methods developed in response to the four \SNAS datasets showcased in open challenges. These challenges have been a driving force in the field, encouraging researchers to continuously improve upon the state of the art.

Overall, this chapter serves as a timestamped summary of the history of the \SNAS datasets and challenges, cataloging over $60$ methods that have been developed or published, many of which include publicly available code to facilitate further research. SoccerNet underscores the profound impact that open-source data, methods, and benchmarks have on advancing the field of video understanding, particularly in the context of sports.

\section*{Acknowledgements}

A. Cioppa is funded by the F.R.S.-FNRS. The research reported in this publication was supported by funding from KAUST Center of Excellence on GenAI, under award number 5940.

%\bibliographystyle{ws-rv-van}
% \bibliography{bib/abbreviation-short, bib/action, bib/activity, bib/dataset, bib/labo, bib/learning, bib/PUT-NEW-REFS-HERE, bib/soccer, bib/soccernet-challenge, bib/sports}

\end{document}